\documentclass{article}
\pdfoutput=1

\PassOptionsToPackage{numbers}{natbib}
\usepackage[dvipsnames,table]{xcolor}

\usepackage[final]{neurips_2023}

\usepackage[utf8]{inputenc} 
\usepackage[T1]{fontenc}    
\usepackage{hyperref}       
\usepackage{url}            
\usepackage{booktabs}       
\usepackage{amsfonts}       
\usepackage{nicefrac}       
\usepackage{microtype}      
\usepackage{xcolor}         

\usepackage[frozencache=true,cachedir=minted-cache]{minted}

\usepackage{url}
\usepackage{booktabs}
\usepackage{amsmath,amsthm,amsfonts,amssymb,amscd}
\usepackage{bbm}
\usepackage{cleveref}
\usepackage{lastpage}
\usepackage{enumerate}
\usepackage{fancyhdr}
\usepackage{mathrsfs}
\usepackage{xcolor}
\usepackage{graphicx}
\usepackage{listings}
\usepackage{mathtools}
\usepackage{algorithm}
\usepackage{algpseudocode}
\usepackage{todonotes}
\usepackage{adjustbox}
\usepackage{wrapfig}
\usepackage{multirow}
\usepackage{makecell}

\newcommand{\KL}{D_{\mathrm{KL}}}

\DeclareMathOperator*{\argmax}{arg\,max}
\DeclareMathOperator*{\argmin}{arg\,min}

\newcommand{\bc}{\mathbf{c}}

\newcommand\Uc{\mathcal{U}}

\newcommand\Rc{\mathcal{R}}
\newcommand\Xc{\mathcal{X}}

\usepackage{stackengine}

\title{Active Preference Inference using Language Models and Probabilistic Reasoning}

\author{%
  Wasu Top Piriyakulkij\\
  Cornell University\\
  \texttt{wp237@cornell.edu}\\
  \And
  Volodymyr Kuleshov\\
  Cornell Tech\\
  \texttt{kuleshov@cornell.edu}\\
  \And
  Kevin Ellis\\
  Cornell University\\
  \texttt{kellis@cornell.edu}\\
}

\begin{document}

\maketitle

\begin{abstract}
Actively inferring user preferences, for example by asking good questions, is important for any human-facing decision-making system.
Active inference allows such systems to adapt and personalize themselves to nuanced individual preferences. To enable this ability for instruction-tuned large language models (LLMs), one may prompt them to ask users questions to infer their preferences, transforming the language models into more robust, interactive systems. However, out of the box, these models are not efficient at extracting preferences: the questions they generate are not informative, requiring a high number of user interactions and impeding the usability of the downstream system. In this work, we introduce an inference-time algorithm that helps LLMs quickly infer preferences by using more informative questions. Our algorithm uses a probabilistic model whose conditional distributions are defined by prompting an LLM, and returns questions that optimize expected entropy and expected model change. Results in a simplified interactive web shopping setting with real product items show that an LLM equipped with our entropy reduction algorithm outperforms baselines with the same underlying LLM on task performance while using fewer user interactions.
\end{abstract}

\section{Introduction}

For any decision-making AI system, knowing each user's task preferences is crucial for the system to be most helpful to individual users. Ideally, in the context of language-based AI systems, we might naively hope for users to fully specify a task in natural language. However, while human speakers can be very effective at selecting their utterances \citep{grice1991studies}, they often fail to fully specify their task preferences to human or AI listeners. Designing language-supported AI systems that can actively infer users' task preferences (e.g., by asking questions when they are uncertain) could help make these technologies more aligned, more personalized, and ultimately more broadly applicable.

Recently, large language models (LLMs) \citep{brown2020gpt, touvron2023llama}, enhanced with instruction tuning \citep{christiano2017rlhf, ouyang2022rlhf}, have shown impressive performance at following natural language task instructions. While LLMs are not natively trained to be good active learners, we can prompt them, leveraging their in-context learning ability, to ask users questions to infer their preferences.
Concurrent work \citep{li2023gate} has shown that prompting LLMs to elicit user preferences indeed helps improve performances over various tasks. Using language models to first elicit user preferences before carrying out the tasks provides a promising path toward enabling these models in domains that require fine-grained understanding of user preferences. 

Nevertheless, a simple conversation with these off-the-shelf LLMs (\Cref{fig:conversation_gpt_4}) reveals their poor ability to efficiently learn user preferences from few questions.
Although these LLMs are able to ask relevant questions when prompted, they do not ask the more informative questions, given the current context. This increases the amount of 
  model-human interaction needed to infer the user's preferences.

We propose an inference-time algorithm for active inference of user preferences by LLMs. 
Our algorithm is motivated by the observation that while out-of-the-box LLMs are suboptimal at interactively inferring user preferences, they are decent at generating diverse questions for extracting preferences, and can generate plausible answers to each such question.
Moreover, given an ongoing interaction with a user, LLMs can reasonably score how likely the user will answer a question in certain ways.
Thus, we can use the LLM-defined scores to define a full probabilistic generative model of what question to ask, how the user might respond, and how those responses update the system's probabilistic beliefs. 
Concretely, our algorithm samples a finite set of questions from the LLM and chooses the one that maximizes information gain.

We evaluate our approach in a simplified web shopping setting using data from a realistic web shopping benchmark~\citep{yao2022webshop}.
Each task consists of a list of products, together with a user who has a latent preference for a specific target product from that list.
LLMs will ask a series of yes/no questions to infer the target product. We compare our inference-time active inference method against a vanilla instruction-tuned LLM \citep{christiano2017rlhf, ouyang2022rlhf} and ReAct \citep{yao2023react}, finding that our approach outperforms both baselines while using fewer questions. In line with recent work \citep{dohan2022cascade, zhao2023lampp, ellis2023human}, our results should be seen as a way of augmenting LLMs with inference-time probabilistic reasoning.

\begin{figure*}[t]
\centering
\includegraphics[width=0.9\textwidth]{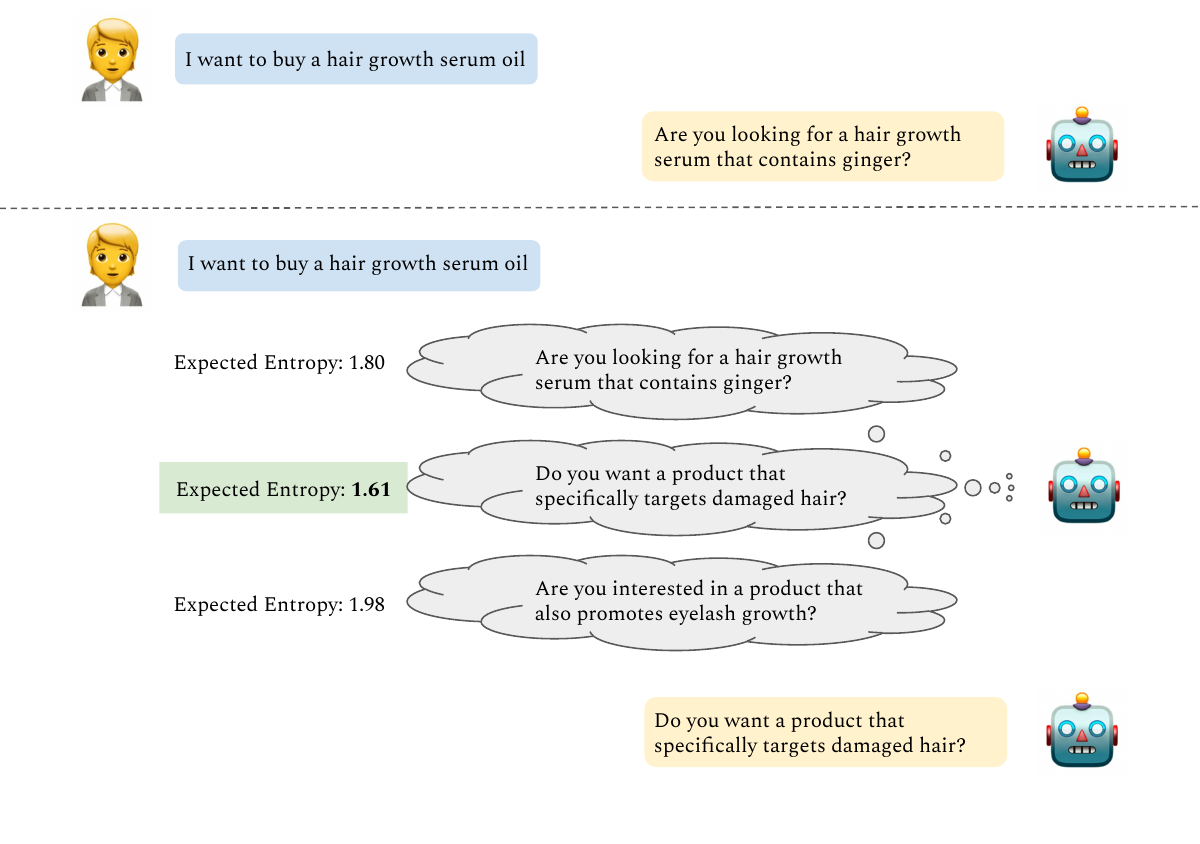}
\caption{(Top) Vanilla instruction-tuned LLM prompted to be a hair growth serum oil seller. (Bottom) LLM with inference-time expected entropy reduction algorithm.}
\vspace{-10pt}
\label{fig:front}
\end{figure*}

\section{Active Preference Inference}

\subsection{Task formulation}

We first introduce the active preference inference formulation. For decision-making tasks, human preferences can be viewed as reward functions that map decisions to a scalar, $R: \Xc \rightarrow \Rc$ where $\Xc$ is the space of decisions. The human user can be described with a reward function and an answer model $(R^*, p_\text{human}(a|q))$ where $q, a \in \Sigma^*$. $q$ and $a$ are a natural language question and answer, and $\Sigma^*$ is the space of natural language. Active preference inference task boils down to a task of asking a series of questions $q_1, .., q_k$, to which the answers $a_i \sim p_\text{human}(a|q_i)$ will be generated, to infer the unknown target reward $R^*$.

Active preference inference resembles active learning but differs in a few key ways.
First, active preference inference separates the answer model $p_{human}(a|q)$ from the reward function $R^*$. This formulation allows the flexibility of asking arbitrary questions to the user. We also note that the question $q$ can be of the form "Is the following example positive or negative: \texttt{\{example\}}", which means active preference inference can be seen as a generalization of active learning. Second, active learning usually assumes access to unlabeled data points that an algorithm can query, while active preference inference asks an algorithm to generate its own questions. 

\subsection{Existing approaches}

Prior to the advent of LLMs, there were attempts at training NLP systems specifically for interactive inference \citep{rao2018learningtoask, yu2020interactive}. Although these works do not consider the question of how to infer the reward function from the final result conversation, we can easily define the predicted reward function as LLM conditioned with that conversation. The main downside of these methods, however, is that they require expensive task-specific data collection for training. Having to gather data for every new task severely limits the usability of these methods.

LLM-based approaches to active preference inference promise to bypass the data collection process with its zero-shot learning capability. With instruction-tuned LLMs, we can directly prompt the models to ask users questions to infer preferences.  However, as mentioned and confirmed later in our experiments, these models turn out to be ineffective at preference inference off-the-shelf and require many interactions with the users before arriving at a good prediction of the reward function.

In this work, we equip LLMs with inference-time probabilistic reasoning algorithm defined in \Cref{sec:main_method} to address this core issue, extracting as much information from the user as possible in as few questions as needed.
This essentially minimizes human-user effort, at the expense of more LLM computation.
As LLMs inference will likely become cheaper in the future, we believe this is the right tradeoff to make.
\Cref{tab:methods} compares other methods that with alternative tradeoffs.

\begin{table}[t]
\adjustbox{max width=\textwidth}{
\centering
\begin{tabular}{cccc}
\toprule
& Task-specific data collection & Informative-question-asking ability & Inference cost\\ 
\midrule
\makecell{Traditional interactive NLP system \\ \citep{rao2018learningtoask, yu2020interactive}} & Required & Optimized during training & Low\\
\makecell{LLM w/o probabilistic reasoning \\ \citep{li2023gate}} & Not required & Not optimized & Medium\\
\makecell{\textbf{LLM w/ probabilistic reasoning} \\ \textbf{(ours)}} & Not required & Optimized during inference & High\\
\bottomrule
\end{tabular}}
\caption{Comparison between different existing approaches for active preference inference}
\vspace{-25pt}
\label{tab:methods}
\end{table}

\section{Inference-time probabilistic reasoning for asking informative question} \label{sec:main_method}

In this section, we define the relevant models and present two measures from the active learning literature \citep{settles2009active} that one may use to pick an informative question.

\subsection{Model definition} \label{method:def}

To develop our model, we make an assumption that the target reward function $R^*$ is a binary function that is 1 only at the target decision $x^* \in \Xc$ and 0 everywhere else. With this assumption, we only need to infer the target decision $x^*$ to know $R^*$.

We define the joint distribution as $p(x, q, a) = p(a | x, q) p(x) p(q)$. The distributions needed to compute the joint are defined by prompting an LLM:
\begin{align*}
    p(q) &= \text{Uniform}(\Sigma^*)\\
    p(x) &= \text{Uniform}(\Xc)\\
    p(a|x, q) &= \; \left[
    \begin{array}{cc}
         &\text{a binary 0/1 score output by prompting an LLM}\\
         &\text{whether the answer $a$ to the question $q$ is consistent}\\
         &\text{with the product description of $x$}
    \end{array}\right]\\
\end{align*}
With these defined distributions, we know that $p(x|q, a) \propto p(a|x, q)$ because $p(x|q) = p(x) = \Uc(X)$. We can also easily calculate $p(a|q) = \sum_{x \in X} p(a|x, q) p(x) $. In addition, we define a question proposal distribution $r(q)$ by asking an LLM to generate a finite number of possible questions given all product information $X$. Prompts used for obtaining $p(a|x, q)$ and $r(q)$ can be founded in \Cref{app:prompts}.

We can easily extend our model to the setting where we have multiple pairs of questions and answers. Let $\bc$ denote a conversation containing a sequence of questions and answers $\bc = \{(q^{(c)}_1, a^{(c)}_1), ..., (q^{(c)}_m, a^{(c)}_m)\}$. Our model becomes $p(x, \bc) = p(x) \prod_i p(a^{(c)}_i|x, q^{(c)}_i) p(q^{(c)}_i)$. With this new model, $p(x|\bc) \propto \prod_i p(a^{(c)}_i|x, q^{(c)}_i)$, and $p(a|\bc, q) \propto \sum_{x \in X} (p(a|x,q) \prod_i p(a^{(c)}_i|x, q^{(c)}_i))$. We also modify the prompt for $r(q)$ to take into account the conversation, notated $r(q|\bc)$. 

\subsection{Objectives for choosing informative questions}

With the distributions defined in \Cref{method:def}, we sample a finite set of questions from the proposal distribution $r(q|\bc)$, calculate the amount of information each question would reveal in expectation, and subsequently pick the question with the highest amount of expected information. We consider two objectives, commonly used in the active learning field \citep{settles2009active}, to measure how informative a question is: expected entropy minimization and expected model change maximization.

\paragraph{Expected Entropy Minimization.} The most informative question could be thought as the question which reduces the most amount of uncertainty in expectation. This notion can be captured with the expected entropy minimization objective. The objective is as follows:
\begin{equation}
    \argmin_{q \sim r(q|\bc)} E_{p(a|\bc, q)} [H(p(x|\bc, q, a)]
\end{equation}

\paragraph{Expected Model Change Maximization.} Another approach is to think about model parameter/belief and consider the most informative question as the question that would make the most change to the model parameter/belief. For our tasks, we can let belief be a decision $x \in \Xc$, representing the notion that we believe $x$ is the decision with a positive reward. We can use KL divergence to measure the change in belief. The objective is as follows:
\begin{equation}
    \argmax_{q \sim r(q|\bc)} E_{p(a|\bc, q)} [\KL(p(x|\bc, q, a)||p(x|\bc))]
\end{equation}

\paragraph{Equivalence between the two objectives.} In our specific setting, it can be shown that expected entropy minimization and expected model change maximization are in fact equivalent. This is due to the fact that the expected cross-entropy $E_{p_\text{model}(a|\bc, q)} [E_{p(h|\bc, q, a)} [-\log p(h|\bc)]$ is the same for all $q, a$ because by our definition, $p(h|\bc, q, a) > 0$ if $p(h|\bc) > 0$, and if $p(h|\bc) > 0$, then $p(h|\bc) = f(\bc)$ is constant to $h$. More details can be founded at \Cref{app:equivalence}

\section{Experiments}

\begin{figure*}[t]
\centering
\includegraphics[width=0.48\textwidth]{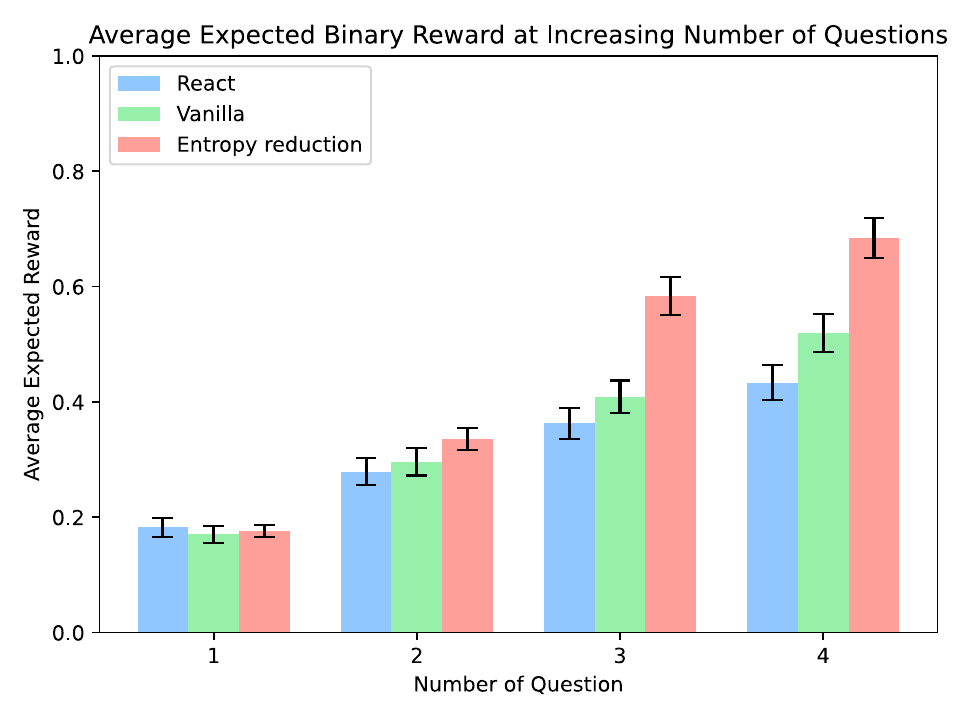}
\includegraphics[width=0.48\textwidth]{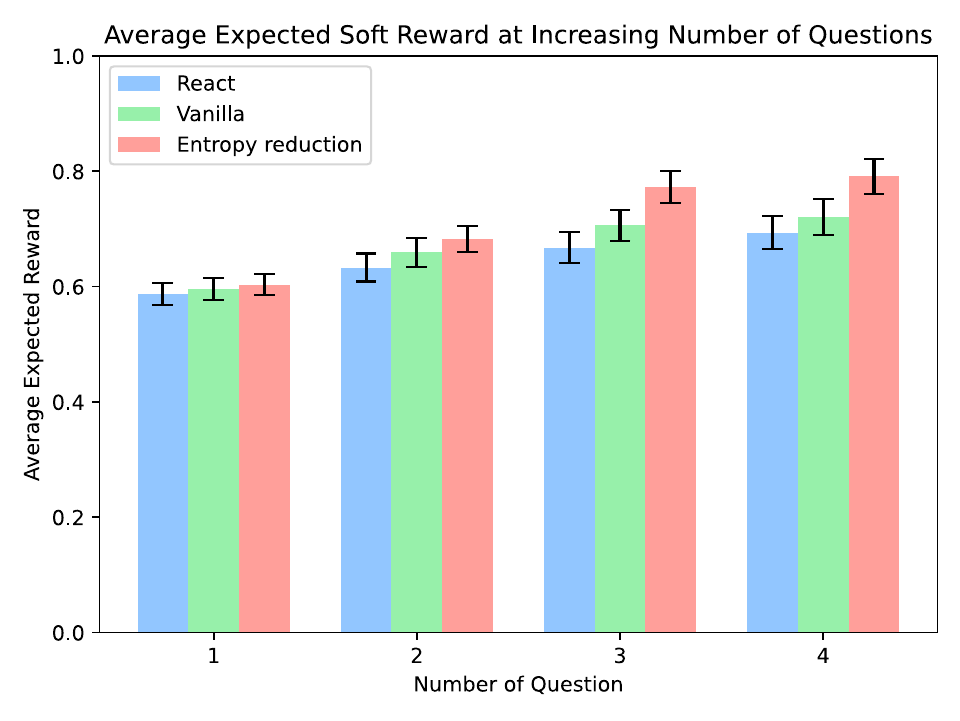}
\caption{(Left) Average expected binary reward at increasing number of number of questions. (Right) Average expected soft reward at increasing number of number of questions.}
\vspace{-10pt}
\label{fig:rewards}
\end{figure*}

We evaluate an LLM augmented with our entropy reduction algorithm (entropy reduction LLM) on 150 simplified web shopping tasks, each task consisting of 10 product items from WebShop \citep{yao2022webshop}. The products in each task come from the same product type, e.g., hair growth serum oil, HDMI cable, phone case, etc. 
While the goal of active preference inference is to infer the target reward function $R^*$, it is more realistic in web-shopping settings to make algorithms return a single product $\hat{x}$ and evaluate them based on the reward of the returned product $R^*(\hat{x})$, so we employ this evaluation in this work. This evaluation can still be seen as implicitly inferring reward function $\hat{R}$ and then return $\hat{x} = \argmax_{x \in \Xc} \hat{R}(x)$. We consider two types of reward functions: binary and soft reward. Both are constructed based on a target product $x^*$ randomly chosen from the task's list of products. More details on task creation can be found at \Cref{app:task-creation}. We simulate human users $p_\text{human}(a|q)$ with GPT-4 \citep{openai2023gpt4} by inputting target product $x^*$ in the prompt. We make a simplifying assumption that only yes/no questions to users are allowed.

Two baselines we consider are vanilla instruction-tuned LLM \citep{christiano2017rlhf, ouyang2022rlhf} and ReAct LLM \citep{yao2023react}. All methods use the same $p(x|q, a)$ to choose a product given questions and answers; the differences in these methods lie in the questions they choose to ask. Vanilla instruction-tuned LLM is prompted to roleplay to be a product seller and ask questions. Our entropy reduction method samples 8 questions from the proposal distribution $r(q|\bc)$ but would reduce down to vanilla instruction-tuned LLM if we only sample one question from the proposal distribution. ReAct is a prompting method for LLM Agents where thinking is part of the action space. In our question-asking scenario, asking a question can be seen as an action. We write a prompt that has a thinking process on what question is best to ask at each point. Prompts for both of these methods can be found at \Cref{app:prompts}. When more than one product has positive probability $p(x|q, a)$ at the end of the questioning, we assume the methods return the positive-probability products at random and actually calculate the expected reward on them. The underlying LLM for all methods is GPT-4. We sample at zero temperature (For the proposal distribution, we actually prompt LLM to return a list of diverse questions instead of sampling questions from LLM with positive temperature).

\subsection{Binary Reward}

\begin{wrapfigure}[19]{r}{0.50\textwidth}
\centering
\includegraphics[width=0.48\textwidth]{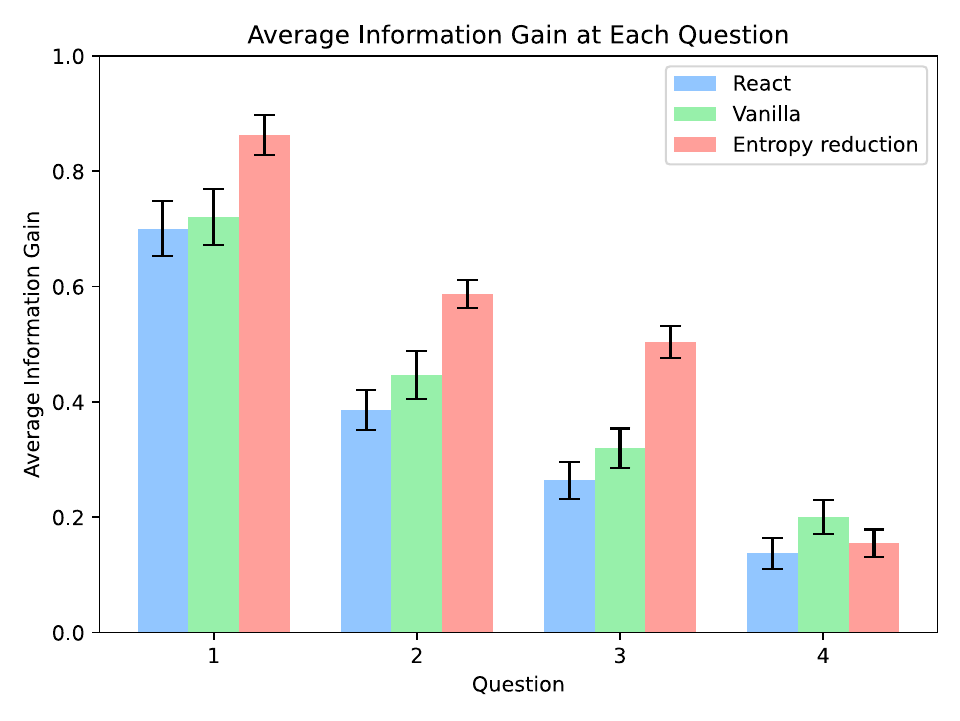}
\caption{Average information gain at each question. Confidence intervals are computed over 150 diffent tasks.}
\label{fig:info-gain}
\end{wrapfigure}

First, we consider a binary reward setting where the reward  is 1 if the target product is retrieved and 0 otherwise. This binary reward is consistent with our modeling assumption.
\Cref{fig:rewards} (left) shows that entropy reduction LLM significantly outperforms both baselines when the number of questions is greater than one. 
In fact, our algorithm after three questions even surpasses the two baselines after four questions.

\Cref{fig:info-gain} reports average information gain at each round of questioning. For the first three questions, we can see that entropy reduction LLM gains more information than both baselines. At the last question, vanilla instruction-tuned LLM gets more information, but all methods do not receive much information gain at this point. There are 10 products in the list of products for each task, and at the fourth question, not much information is left to be extracted from human users.

\subsection{Soft Reward}

We also consider a more realistic, soft reward setting. In practical scenarios, it is likely that user preferences for product items are not all-or-nothing, and user satisfaction upon receiving each product item as output is more fine-grained. We adopt the reward used in the Webshop benchmark \citep{yao2022webshop}. We exclude product price and options from the original reward calculation as they are not relevant in our task setting. The reward is then calculated based on Webshop's text matching heuristics and the attribute overlap between the target product and the returned product. The reward falls within the range of 0 to 1, with an exact match yielding a reward of 1.

Similar to the binary reward, \Cref{fig:rewards} (right) shows that entropy reduction LLM outperforms both baselines when the number of questions is greater than one. Even though the soft reward is different from our modeling assumption, the inference-time entropy reduction algorithm is still better at preference inference than the two baselines. 

\section{Related Work}

\paragraph{Learning to ask clarifying question.} There are many prior works in learning to ask clarification questions \citep{rao2018learningtoask, aliannejadi2019asking, yu2020interactive, zamani2020generating}. \cite{rao2018learningtoask} relies on the idea that a good clarification question is the one whose answer will be useful. They use Q\&A forum data to train neural networks to be an answer model and an answer utility calculator and choose question that maximizes expected utility. \cite{aliannejadi2019asking} looks at the problem of asking good clarifying questions from an information retrieval perspective, and train models with question-answer pairs data to retrieve good questions from a crowdsourced dataset.
\cite{yu2020interactive} studies interactive classification where questions are selected based on the entropy minimization principle similar to our work. They crowdsource non-interactive data to train neural networks used to define their probability distributions. The main difference between our work and these work is that we do not require any sort of training data and instead rely on the powerful zero-shot learning capability of LLMs to define rich distributions, which we can probabilistically reason over at inference time.

\paragraph{Probabilistic reasoning with LLMs.} \cite{dohan2022cascade} provides a unifying framework of seeing recent inference-time LLM algorithms \citep{nye2021scratchpad, wei2022cot, creswell2022selection} as performing reasoning with probabilistic programs. \cite{zhao2023lampp} utilizes LLM to define prior distributions in a probabilistic graphical model and performs inference in the model to solve navigation, action recognition, and image segmentation. Our work studies inference-time probabilistic reasoning with LLM in the context of active preference inference.

\section{Conclusion and Discussion}

In this work, we present an inference-time algorithm to enhance the efficiency of actively inferring preferences of LLMs and demonstrate that in our web shopping setting, our method performs better than the baselines using fewer user interactions.

There are several future directions for this work. The most obvious one is to let the allowed questions be open-ended. Even though this modification will not call for any change to the model definition, this generalization will necessitate an approximation to the objective calculation as there might be infinitely many answers to an open-ended question. One needs to test empirically how much computation is needed before the Monte Carlo approximation is close enough to the true expected value. Another direction is to come up with a more flexible probabilistic model that does not assume the target reward function is binary. While this assumption works well in the web shopping setting, as evidently shown in both binary and soft reward evaluation, it may result in poor performance for other tasks that involve a full prediction of the reward function.

\bibliographystyle{iclr2023_conference}
\bibliography{iclr2023_conference}
\nocite{*}

\appendix
\section{Appendix}

\begin{figure*}[!h]
\centering
\begin{minted}[breaklines=true]{text}
User: Consider the following products:
1. Product 1 - blue, plastic, heavy, dust-proof, iPhone phone case
2. Product 2 - green, plastic, heavy, dust-proof, iPhone phone case
3. Product 3 - purple, plastic, heavy, dust-proof, iPhone phone case
4. Product 4 - green, leather, light, water-proof, iPhone phone case
5. Product 5 - red, plastic, heavy, dust-proof, iPhone phone case
6. Product 6 - red, plastic, heavy, dust-proof, android phone case

Please ask yes or no question to figure which product I want, one at a time. You will only get to ask three questions, so choose wisely. If you're ready, you can go ahead and ask the first question

GPT-4: Is the product you want made of plastic?

User: yes

GPT-4: Is the product you want for an iPhone?

User: yes

GPT-4: Is the product you want blue?

User: no

GPT-4: Is the product you want green?

User: no

GPT-4: Is the product you want red?

User: no

GPT-4: So, the product you want is the purple, plastic, heavy, dust-proof, iPhone phone case. Is that correct?
\end{minted}
\caption{An example conversation with GPT-4 (zero temperature) demonstrating how GPT-4 is lacking efficiency when prompted to ask questions. In this example, one can always figure out the target product within 3 questions, but GPT-4 is using 5 questions, which is as bad as asking if the target product is Product 1 or 2 or 3 and so on.}
\label{fig:conversation_gpt_4}
\end{figure*}

\subsection{Equivalence between expected entropy minimization and expected model change maximization}\label{app:equivalence}

Expected entropy minimization and expected model change maximization can be shown to be equivalent for the setting in this work. The proof is straightforward:
\begin{align*}
    &\argmax_{q \sim r(q|\bc)} E_{p_\text{model}(a|\bc, q)} [\KL(p(h|\bc, q, a)||p(h|\bc))] \\
    &\;= \argmax_{q \sim r(q|\bc)} E_{p_\text{model}(a|\bc, q)} [E_{p(h|\bc, q, a)} [\log p(h|\bc)] - H(p(h|\bc, q, a))]\\
    &\;= \argmax_{q \sim r(q|\bc)} -H(p(h|\bc, q, a))\\
    &\;= \argmin_{q \sim r(q|\bc)} H(p(x|\bc, q, a))
\end{align*}

\subsection{Task Creation}\label{app:task-creation}

To create the tasks we evaluate in our work, we first use GPT-4 to extract product types/categories from actual products from WebShop \citep{yao2022webshop} (see \Cref{fig:prompt_product_extract}). Each task has its own product type, and the list of products for the task is retrieved by searching WebShop website for the top 10 products with the task's product type as the search keyword. The target product is sampled randomly from the list of products. 

\begin{figure*}[!h]
\centering
\begin{minted}[breaklines]{text}
Given a product, please tell me what's the type of the product.

For example, 
Product: Bright Citrus Deodorant by Earth Mama | Natural and Safe for Sensitive Skin, Pregnancy and Breastfeeding, Contains Organic Calendula 3-Ounce
Product type: Deodorant

Here's the product,
Product: {product}

Please put your answer in the format 'Product type: '. Do not say anything else.
\end{minted}
\caption{Prompt used for extracting product type}
\label{fig:prompt_product_extract}
\end{figure*}

\subsection{Prompts}\label{app:prompts}
\Cref{fig:prompt_proposal}, \Cref{fig:prompt_human}, \Cref{fig:prompt_vanilla}, \Cref{fig:prompt_react},
\Cref{fig:prompt_soft_reward} contain the prompts for $r(q|\bc)$, $p_\text{human}(a|q)$, vanilla baseline, ReAct baseline, and soft reward calculation respectively. For $p(x|q, a)$, we actually use the same prompt as $p_\text{human}(a|q)$ to calculate it by simulating a human that prefers $x$ most and ask question $q$ to the simulated human. Suppose $a^{(\text{human})}$ is the answer that the simulated human outputs. The binary score that $p(x|q, a)$ is proportional to is 1 when the answer matches human answer and 0 when it is not.

\begin{figure*}[!h]
\centering
\begin{minted}[breaklines]{text}
Suppose you are a seller, and you need to determine which of the following products the customer wants to buy:
{products}
The conversation you've had with the customer so far is as follows: \n{conversation}\n. \
Generate a list of 10 yes/no questions you would ask the customer to further figure out which of the products the customer want.
Keep in mind that:
1. The question must ask for a yes or no answer only. 
2. The question must ask about one specific thing, do not use 'and' or 'or' to put many features/properties/attributes together.
3. Do not explain.
\end{minted}
\caption{Prompt used for $r(q|\bc)$}
\label{fig:prompt_proposal}
\end{figure*}
\begin{figure*}[!h]
\centering
\begin{minted}[breaklines]{text}
Suppose you want to buy the following product:

{product_txt}

You will interact with a salesperson. They will ask you a question about the product you want. Please answer only either yes or no based on the product information.
If you are asked about features not mentioned in the product description, then say No.
\end{minted}
\caption{Prompt used for simulating $p_\text{human}(a|q)$ and also for $p(x|q,a)$}
\label{fig:prompt_human}
\end{figure*}
\begin{figure*}[!h]
\centering
\begin{minted}[breaklines]{text}
Suppose you are a seller, and you need to determine which of the following products the customer wants to buy:
{products}
The conversation you've had with the customer so far is as follows: \n{conversation}\n. \
What's the next yes/no question you would ask the customer to further figure out which of the products the customer want?
Keep in mind that:
1. The question must ask for a yes or no answer only. 
2. The question must ask about one specific thing, do not use 'and' or 'or' to put many features/properties/attributes together.
3. Do not explain.
\end{minted}
\caption{Prompt used for roleplaying Vanilla LLM}
\label{fig:prompt_vanilla}
\end{figure*}
\begin{figure*}[!h]
\centering
\begin{minted}[breaklines]{text}
Please choose the appropriate action (think, ask_question, get_products, choose_products) based on the observation and given example runs.

Important: the question must be a yes/no question

Example 1:

Instruction: i want to buy a phone case

Action: show_products[]
Observation: 
1. ID: B09G9D18YS | Product Name: BURGA Phone Case Compatible with iPhone 13 - Hybrid 2-Layer Hard Shell + Silicone Protective Case - Black Polka Dots Pattern Nude Almond Latte Fashion - Scratch-Resistant Shockproof Cover
- Price: $19.95
- Attributes: phone case, wireless charging, heavy duty
- Options: color (almond latte, black & gold onyx, dazzling glow, emerald pool, fatal contradiction, gentle wind, gold dust, hidden beauty, iconic ruby, mystic river)

... # Full list of products in the actual prompt not included here for brevity

Action: think[I'll ask a few yes/no questions to determine which one of these products the customer wants. Some of these phone cases are holster phone cases. I'll ask a yes/no question about if they want a holster]
Observation: OK.

Action: ask_question[Are you looking for a holster phone case? (Yes/No)]
Observation: Answer: No.

Action: think[Ok, we're looking for non-holster phone case. Among these, there are some clear phone cases. Let's ask a yes/no question if they want the phone case to be clear]
Observation: OK.

Action: ask_question[Are you looking for a clear phone case? (Yes/No)]
Observation: Answer: No.

... # Full question asking in the actual prompt not included for brevity

IMPORTANT: No more question can be asked.

Action: choose_product[]
Observation: OK.

Example 2:

Instruction: {instruction}
\end{minted}
\caption{Prompt used for ReAct}
\label{fig:prompt_react}
\end{figure*}
\begin{figure*}[!h]
\centering
\begin{minted}[breaklines]{text}
Given {product_type} with the following description:

{target_product_txt}

Please rate each of the following {product_type} on a scale of 1-10 based on how similar it is to the given target {product_type}

{products_txt}

The {product_type} least similar to the given target {product_type} should receive a score of 1. Please give you answer in the format '1. Explanation: one sentence, Rating: x/10' and so on. Always explain and give rating to all answers. At the very end, please put all output ratings in the form 'All ratings: score_1, score_2, ...
\end{minted}
\caption{Prompt used to calculate soft reward}
\label{fig:prompt_soft_reward}
\end{figure*}

\end{document}